\documentclass[10pt,conference,Letterpaper]{IEEEtran}
\IEEEoverridecommandlockouts
\usepackage{CJKutf8}
\usepackage[hidelinks]{hyperref}
\usepackage[cmex10]{amsmath}
\usepackage{amssymb,amsfonts}
\usepackage{dblfloatfix}
\usepackage[ruled,vlined]{algorithm2e}
\usepackage{graphicx}
\graphicspath{{Figures/PDF/}{Figures/PNG/}}
\usepackage{ragged2e} 
\usepackage{booktabs,makecell, multirow, tabularx}
\usepackage{booktabs}
\usepackage{balance}
\usepackage{siunitx}
\usepackage[numbers,compress]{natbib}
\usepackage{texnames}
\usepackage{bm,bbm}
\usepackage{orcidlink}
\usepackage{float}
\usepackage{color}
\usepackage{colortbl}
\newcommand\crule[3][black]{\textcolor{#1}{\rule{#2}{#3}}}
\def\clegend#1#2{\crule[#1]{6pt}{6pt} {\color{#1} #2}}
\definecolor{Lwater}{RGB}{0, 112, 255}
\definecolor{Lroad}{RGB}{197, 0, 255}
\definecolor{Lbuildings}{RGB}{255, 255, 0}
\definecolor{Lfarmland}{RGB}{85, 255, 0}
\definecolor{Lforest}{RGB}{0, 115, 76}
\definecolor{Lbareland}{RGB}{255, 235, 190}
\definecolor{Limpervioussurface}{RGB}{255, 170, 0}
\makeatletter
\newcommand{\linebreakand}{%
  \end{@IEEEauthorhalign}
  \hfill\mbox{}\par
  \mbox{}\hfill\begin{@IEEEauthorhalign}
}
\makeatother

\begin{document}
\begin{CJK}{UTF8}{gbsn}

\title{DEPTHSEG: DEPTH PROMPTING IN REMOTE SENSING SEMANTIC SEGMENTATION
\thanks{This work was supported by the National Natural Science Foundation of China under grant 42401451 and 42271416, the National Key R\&D Program of China under grant 2023YFD2201702, the Hubei Natural Science Foundation under grant 2024AFB223, and the Guangxi Science and Technology Major Project under grant AA22068072.}}

\author{\IEEEauthorblockN{Ning Zhou\orcidlink{0000-0002-3967-4032}}
	\IEEEauthorblockA{\textit{Wuhan University}\\
		430079 Wuhan, China\\
		zhouning@whu.edu.cn}
	\and
	\IEEEauthorblockN{Shanxiong Chen}
	\IEEEauthorblockA{\textit{Changjiang Spatial Information Technology Engineering Co., Ltd}\\
		430010 Wuhan, China\\
		shanxiongchen@whu.edu.cn}
	\and
	\IEEEauthorblockN{Mingting Zhou\orcidlink{0000-0002-5150-4511}}
	\IEEEauthorblockA{\textit{Wuhan University}\\
		430079 Wuhan, China\\
		mintyzhou@whu.edu.cn}
	\linebreakand 
	\IEEEauthorblockN{Haigang Sui}
	\IEEEauthorblockA{\textit{Wuhan University}\\
		430079 Wuhan, China\\
		haigang\underline{~}sui@263.net}
	\and
	\IEEEauthorblockN{Lieyun Hu}
	\IEEEauthorblockA{\textit{Wuhan University}\\
		430079 Wuhan, China\\
		lieyun\underline{~}hu@whu.edu.cn}
	\and
	\IEEEauthorblockN{Han Li}
	\IEEEauthorblockA{\textit{Wuhan University}\\
		430079 Wuhan, China\\
		xmy\underline{~}soul@whu.edu.cn}
	\and
	\IEEEauthorblockN{Li Hua}
	\IEEEauthorblockA{\textit{Huazhong Agr. University}\\
		430070 Wuhan, China\\
		huali@mail.hzau.edu.cn}
	\and
	\IEEEauthorblockN{Qiming Zhou\orcidlink{0000-0003-0934-0602}}
	\IEEEauthorblockA{\textit{Hong Kong Baptist University}\\
		999077 Hong Kong, China\\
		qiming@hkbu.edu.hk}
}
\maketitle
\begin{abstract}
Remote sensing semantic segmentation is crucial for extracting detailed land surface information, enabling applications such as environmental monitoring, land use planning, and resource assessment. In recent years, advancements in artificial intelligence have spurred the development of automatic remote sensing semantic segmentation methods. However, the existing semantic segmentation methods focus on distinguishing spectral characteristics of different objects while ignoring the differences in the elevation of the different targets. This results in land cover misclassification in complex scenarios involving shadow occlusion and spectral confusion. In this paper, we introduce a depth prompting two-dimensional (2D) remote sensing semantic segmentation framework (DepthSeg). It automatically models depth/height information from 2D remote sensing images and integrates it into the semantic segmentation framework to mitigate the effects of spectral confusion and shadow occlusion. During the feature extraction phase of DepthSeg, we introduce a lightweight adapter to enable cost-effective fine-tuning of the large-parameter vision transformer encoder pre-trained by natural images. In the depth prompting phase, we propose a depth prompter to model depth/height features explicitly. In the semantic prediction phase, we introduce a semantic classification decoder that couples the depth prompts with high-dimensional land-cover features, enabling accurate extraction of land-cover types. Experiments on the LiuZhou dataset validate the advantages of the DepthSeg framework in land cover mapping tasks. Detailed ablation studies further highlight the significance of the depth prompts in remote sensing semantic segmentation. 
\end{abstract}

\begin{IEEEkeywords}
	Depth prompt, semantic segmentation, lightweight adapter.
\end{IEEEkeywords}

\section{Introduction}

Remote sensing semantic segmentation aims to classify each pixel in satellite or aerial imagery into specific land cover types, playing a vital role in natural resource monitoring, urban planning, and environmental conservation. Current mainstream methods for semantic segmentation in remote sensing images predominantly rely on data-driven deep learning approaches\citep{long2015fully,ronneberger2015u,badrinarayanan2017segnet,chen2017deeplab,chen2018encoder,zhao2017pyramid,xie2021segformer,ST-UNet,CTCFNet}. Techniques such as Fully Convolutional Networks\citep{long2015fully}, U-shaped Networks\citep{ronneberger2015u}, Segmentation Networks\citep{badrinarayanan2017segnet}, DeepLab\citep{chen2017deeplab}, and Segmentation Transformers\citep{xie2021segformer} have advanced the field by eliminating the need for manual feature extraction through the use of meticulously designed trainable modules. These methods have improved segmentation accuracy and made the process more intelligent. However, their accuracy often decreases when applied to complex environments. This limitation is primarily due to the challenging imaging conditions of remote sensing and the complex characteristics of the Earth's surface. Remote sensing images can exhibit spectral similarities or geometric similarities between objects. In addition, non-orthographic imaging introduces numerous shadows in remote sensing images, making it challenging to identify land cover classes in shadowed areas. These factors collectively lead to land cover misclassification with existing semantic segmentation methods, limiting their potential in real-world applications.

In real-world scenarios, the different objects not only exhibit distinct geometric and spectral characteristics but also often differ in their elevation or depth. Incorporating three-dimensional (3D) information facilitates distinguishing between objects of similar spectral characteristics but differing depths or heights, such as building rooftops and plazas with similar materials. Moreover, leveraging depth information helps mitigate the land cover misclassification caused by challenges such as shadow occlusion. In remote sensing, there has been research utilizing height information, relying on stereo imagery to derive Earth's surface elevation\citep{NI2015,YU2023,CHEN2023113802}. However, the high cost of acquiring stereo imagery poses challenges for large-scale applications in land-cover mapping. Inferring elevation from 2D remote sensing imagery to enhance semantic segmentation would bring new hope for achieving high-precision and extensive land-cover mapping.  

This paper presents a novel depth prompting remote sensing semantic segmentation framework named DepthSeg to improve the accuracy in land-cover mapping. The proposed DepthSeg framework infers depth information from 2D satellite images and takes the estimated depth as a prompt to reduce the land-cover misclassification caused by spectral confusion and shadow occlusion. A lightweight adapter built on a pre-trained vision transformer (ViT) is taken as the encoder of DepthSeg to capture the land-cover features in 2D remote sensing imagery with a low-cost fine-tuning workload. A depth prompter is then introduced to model depth/height features explicitly. The elevation information is integrated into the land-cover mapping process to overcome the effects of shadows and enhance the model's ability to distinguish between spectrally similar objects. Finally, a semantic classifier that integrates the depth prompts and multi-scale object features is introduced to interpret the types of land cover.

\section{Methodology}

\begin{figure}[H]
  \centering
  \setlength{\abovecaptionskip}{-0.5cm}
  \includegraphics[width=1\linewidth]{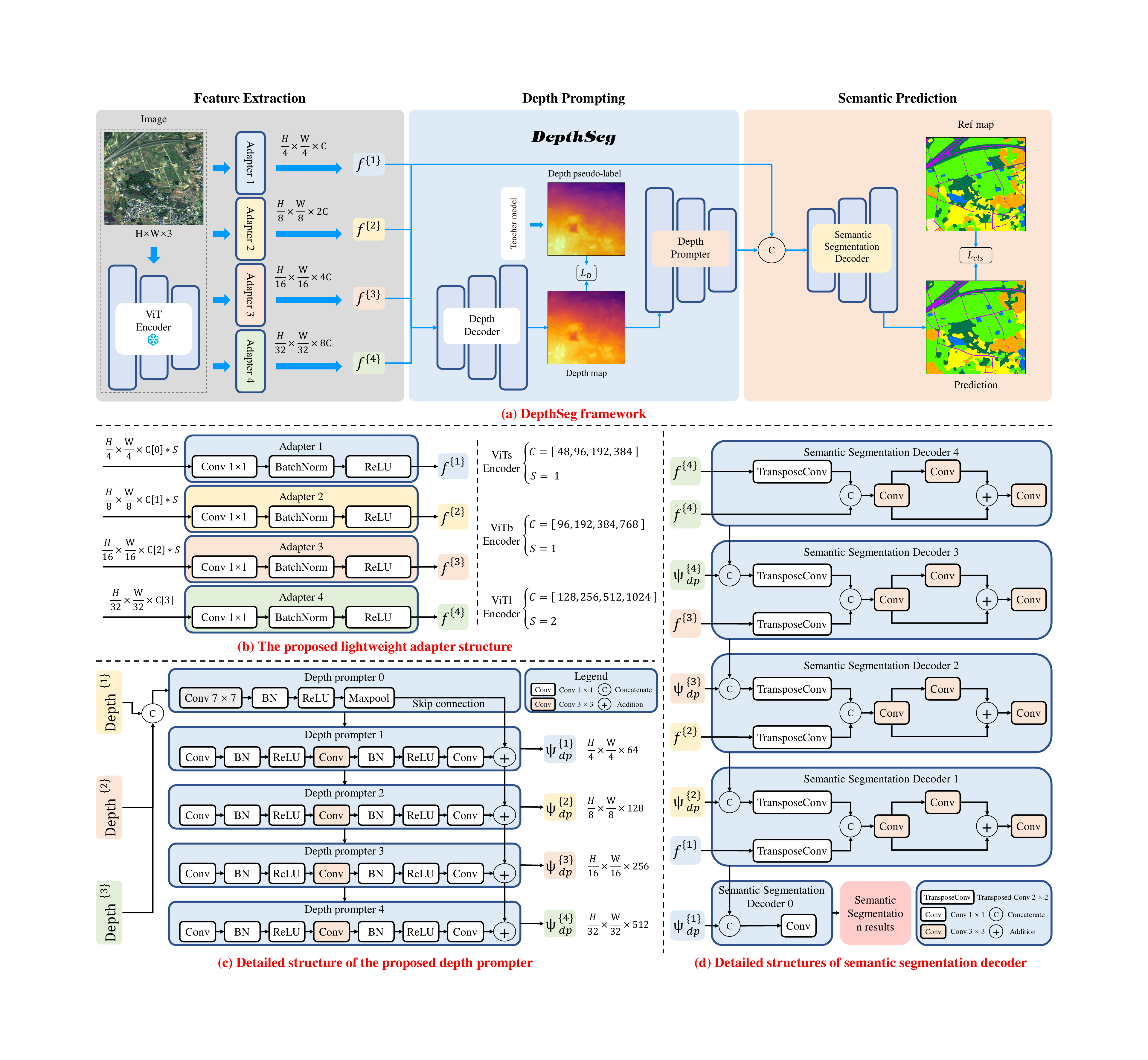}
  \caption{Overview of the proposed depth prompting 2D remote sensing semantic segmentation framework. In (a), the different colors of $f$ denote the high-dimensional image features, while $L$ refers to the loss functions. In (b), $C$ represents the number of channels, and $S$ is the harmonic coefficient for the number of feature channels at different scales. In (c), the three colors of $Depth$ represent the three scales of depth maps output by the dense prediction transformer and the different colors of $\psi$ represent depth prompts.}
  \label{fig:DepthSeg}
\end{figure}

A depth prompting remote sensing semantic segmentation framework named DepthSeg is proposed in this paper to reduce the land-cover misclassification in complex scenarios caused by factors such as shadow occlusion and spectral confusion. The core idea of DepthSeg is to infer depth information from 2D remote sensing images and then embed depth information into the semantic segmentation framework. The DepthSeg framework is illustrated in Fig.~\ref{fig:DepthSeg}. 

\vspace{-0.4cm}
\vspace{-0.3cm}

DepthSeg comprises a feature extraction stage, a depth prompting stage, and a semantic prediction stage. During the feature extraction stage, a frozen ViT encoder\citep{oquab2024dinov2learningrobustvisual} is employed to extract features. To efficiently adapt the features learned from natural images for remote sensing imagery, a lightweight adapter is introduced to fine-tune the pre-trained ViT encoder at a low cost. The depth prompting stage extracts the depth features of objects from 2D remote sensing images. Specifically, a dense prediction transformer\citep{dpt} is introduced to extract depth information, and a depth prompter is introduced to encode depth features, effectively guiding the semantic segmentation. Finally, a semantic segmentation decoder is introduced in the semantic prediction stage to accurately extract the classification of land cover by coupling depth information with high-dimensional land-cover features.

\subsection{Lightweight adapter}

To achieve cost-effective fine-tuning of the pre-trained ViT encoder, a lightweight adapter is designed to facilitate knowledge transfer from natural images to remote sensing images (Fig.\ref{fig:DepthSeg}(b)). The lightweight adapter receives as input the features from the ViT encoder at four different scales, and the dimensions and channel numbers of the features remain unchanged after passing through the adapter. The adapter consists of four convolutional blocks, with each block containing a 1 × 1 convolutional layer, a batch normalization layer, and a rectified linear unit (ReLU) layer. The number of output feature channels from both the ViT encoder and the adapter is related to the size of the ViT. The number of parameters in the adapter also increases with the input feature channel count; however, compared to the ViT encoder, the parameter count is significantly reduced, effectively lowering the training cost.

\subsection{Depth prompter}

The depth prompter is proposed to accurately extract the depth information from 2D remote sensing images and provide prompts for the semantic segmentation (Fig.\ref{fig:DepthSeg}(c)). Embedding depth prompts is an effective approach to address the land cover misclassification caused by factors such as spectral confusion and shadow occlusion. The input to the depth prompter comprises the three scales of shallow depth features. These depth features are processed through the five submodules of the depth prompter to yield high-dimensional depth prompts. The depth prompter consists of a shallow convolutional block for downsampling and four deep encoding convolutional blocks. Feature encoding among the five submodules is facilitated through skip connections and layer-by-layer transmission in a residual-like manner, preserving larger-scale features while extracting deeper-level depth prompts.

\subsection{Semantic segmentation decoder}

Unlike conventional semantic segmentation decoders that only utilize features extracted from bi-temporal image encoders, the proposed semantic segmentation decoder jointly leverages depth prompts $\psi$ and image features $f$ to decode land-cover types (Fig.~\ref{fig:DepthSeg}(d)). The semantic segmentation decoder consists of an input layer, three intermediate layers, and an output layer. The decoded land-cover semantic features are passed layer by layer, ultimately producing the land-cover classification results. 

\subsection{Loss function}

Two loss functions are designed to supervise the DepthSeg framework. To supervise the depth decoder, the loss $L_D$ is designed by integrating the average depth, depth standard deviation, and depth structural features. In addition, $L_{cls}$ is formulated based on cross-entropy loss to supervise the semantic segmentation decoder. Finally, the overall loss functions $L$ for the DepthSeg framework are computed through a combination of these loss components. During the semi-supervised training of the depth decoder, the objective is to align the predicted depth map $X$ with the pseudo-label $Y$ generated by the teacher model. This alignment is achieved by optimizing the depth decoder. The definition of $\mathcal{L}_{D(X,Y)}$ is based on the structure similarity index measure (SSIM)\citep{SSIM} ($\mathcal{L}_{D(X,Y)}=1-SSIM(X,Y)$). Furthermore, during the fully supervised training of the decoder, cross-entropy loss is employed as the loss function for land-cover classification. The definition is $\mathcal{L}_{\mathrm{cls}}=-[Y_{gt}\log\bigl(Y_{pred}\bigr)+(1-Y_{gt})\log\bigl(1-Y_{pred}\bigr)\bigr]$, where $Y_{gt}$ is the label and $Y_{pred}$ is the prediction. In summary, the definition of the total loss function for DepthSeg is $\mathcal{L}=\mathcal{L}_{\mathrm{D(X,Y)}}+\mathcal{L}_{\mathrm{cls}}$.

\section{Experiments and Results}

\subsection{Experimental settings}

\subsubsection{Study area and data}
\begin{figure}[H]
  \centering
  \setlength{\abovecaptionskip}{-0.5cm}
  \includegraphics[width=1\linewidth]{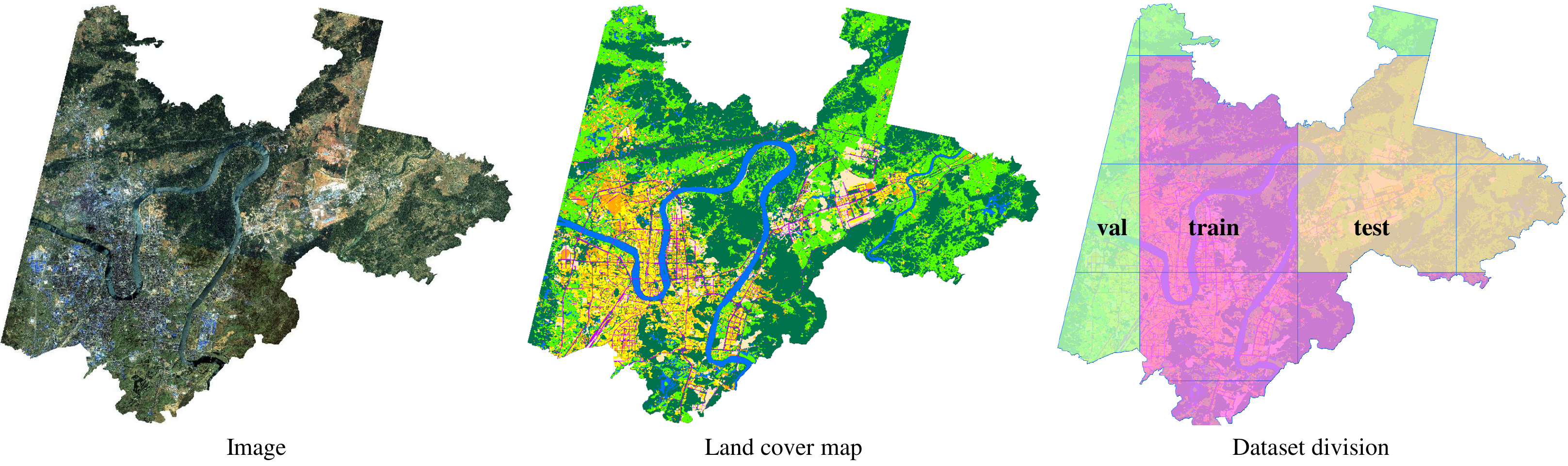}
  \caption{The LiuZhou dataset.}
  \label{fig:dataset}
\end{figure}

To comprehensively evaluate the performance of DepthSeg in real-world application scenarios, we constructed a large-scale land-cover semantic segmentation dataset named LiuZhou. The LiuZhou dataset consists of fused multispectral and panchromatic images from GaoFen-2, acquired in 2015, with a spatial resolution of 0.8 m. It captures land-cover characteristics of a region in Liuzhou, Guangxi, China. The dataset includes seven land-cover categories: cropland, forest, buildings, roads, impervious surfaces, bare land, and water bodies. Using the acquired imagery and professional GIS software, all images were annotated by expert remote sensing specialists. The dataset is divided into training, validation, and the testing subsets, as shown in Fig.\ref{fig:dataset}. The training set consists of 7,047 pairs of 512 × 512 samples, the validation sets include 4,209 pairs, and testing sets contain 4,071 pairs. To ensure the validity of performance evaluation, the training, validation, and testing data are spatially non-overlapping. 

\vspace{-0.4cm}
\vspace{-0.4cm}

\subsubsection{Implementation details and comparative methods}

The DepthSeg framework was implemented in PyTorch and was trained on an NVIDIA RTX 4080 GPU with 16 GB of memory. Depending on the layers and parameters of the ViT encoder, the proposed DepthSeg architectures are available in small, base, and large versions. We utilized the pre-trained encoder provided by \citep{oquab2024dinov2learningrobustvisual}. To effectively supervise the output of the depth decoder, the Depth Anything\citep{depthanything} model was used as a teacher model for the semi-supervised training. During training, the parameters of the encoder were frozen, while the parameters of the other modules in the DepthSeg framework were optimized using the AdamW optimizer with a weight decay of 0.001 and a momentum of 0.9. The initial learning rate was set to 0.0001, and the learning rate schedule followed a MultiStepLR strategy, reducing the learning rate to 0.2 times the current value at 30\% and 60\% of the training steps. The total number of training epochs was set to 50, with the batch size adjusted according to the encoder size: 8 for ViT-s, 4 for ViT-b, and 2 for ViT-l. 

To provide a more objective assessment of the effectiveness of the proposed DepthSeg framework, three methods were selected for comparison: Deeplabv3\citep{chen2017rethinking}, ST-UNet\citep{ST-UNet}, CTCFNet\citep{CTCFNet}. For all the comparative methods, we obtained the publicly available code from the authors and implemented them using the hyperparameters recommended in the original papers, conducting a rigorous comparison across the LiuZhou datasets.

\subsubsection{Evaluation metrics}

Six commonly used evaluation metrics are utilized here for quantifying the accuracy in semantic segmentation. These metrics are the mean precision: $\mathrm{mPre}=\frac1N\sum_{i=1}^N\frac{\mathrm{TP}}{\mathrm{TP}+\mathrm{FP}}$, mean recall: $\mathrm{mRecall}=\frac1N\sum_{i=1}^N\frac{\mathrm{TP}}{\mathrm{TP}+\mathrm{FN}}$, mean F1-score: $\mathrm{mF1}=\frac1N\sum_{i=1}^N\frac{\mathrm{2TP}}{\mathrm{2TP}+\mathrm{FP}+\mathrm{FN}}$, mean intersection over union: $\mathrm{mIoU}=\frac1N\sum_{i=1}^N\frac{\mathrm{TP}}{\mathrm{TP}+\mathrm{FP}+\mathrm{FN}}$, Cohen's Kappa coefficient: $\mathrm{Kappa}=\frac{2(\text{TP×TN}-\text{FN×FP})}{(\text{TP}+\text{FP})(\text{FP}+\text{TN})+(\text{TP}+\text{FN})(\text{FN}+\text{TN})}$, and overall accuracy: $\mathrm{OA}=\frac1N\sum_{i=1}^N\frac{\mathrm{TP}+\mathrm{TN}}{\mathrm{TP}+\mathrm{TN}+\mathrm{FP}+\mathrm{FN}}$. In these formulas, TP are true positives, FP are false positives, FN are false negatives, and the N is the number of land cover classes.
 
\subsection{Results}

\begin{figure}[ht]
  \centering
  \setlength{\abovecaptionskip}{-0.5cm}
  \includegraphics[width=1\linewidth]{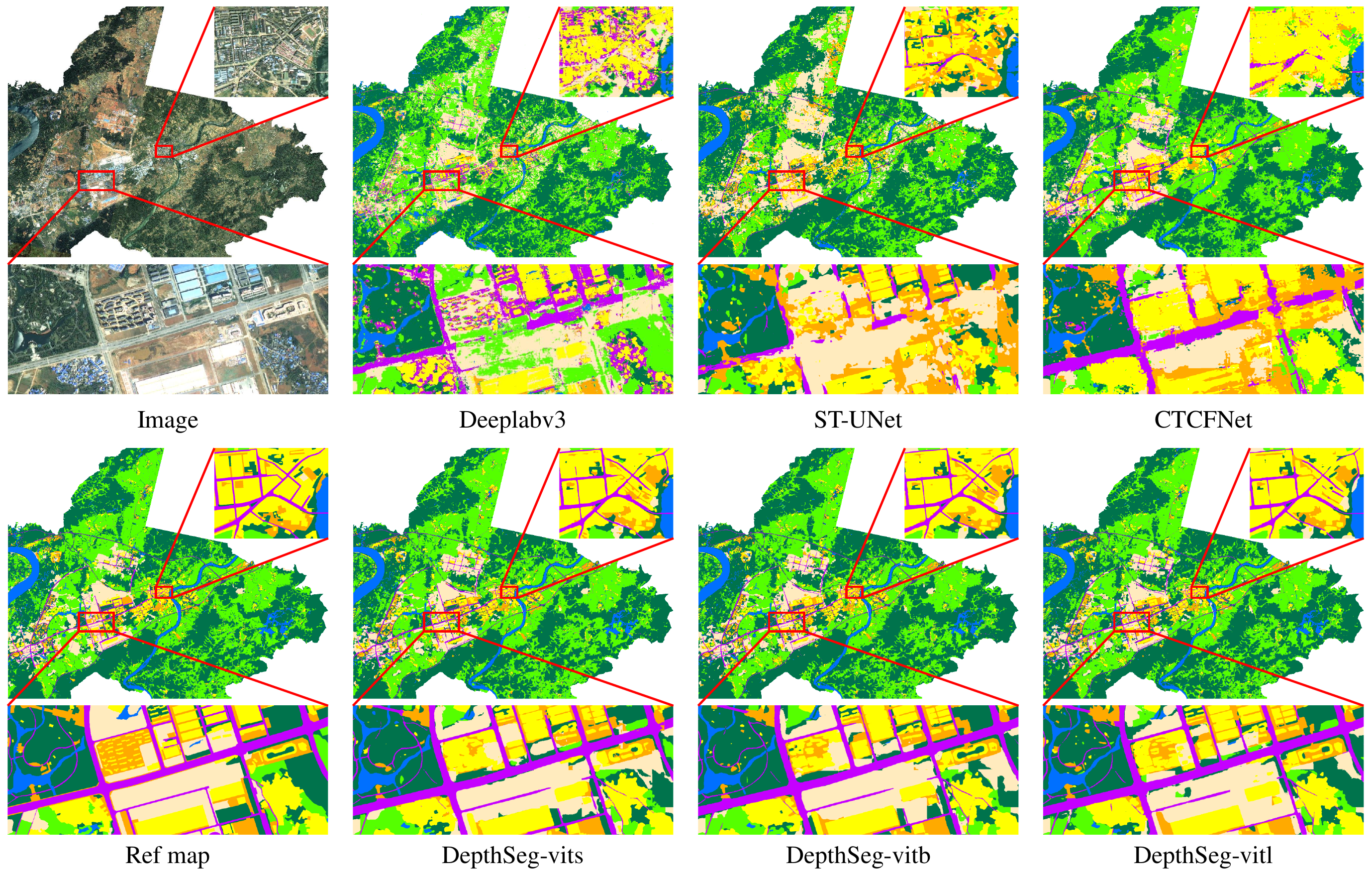}
  \caption{Visualization of the semantic segmentation results of the different methods on the LiuZhou dataset. 
  Legend:
      \clegend{Lwater}{Water},
      \clegend{Lroad}{Road},
      \clegend{Lbuildings}{Buildings},
      \clegend{Lfarmland}{Farmland},
      \clegend{Lforest}{Forest},
      \clegend{Lbareland}{Bare land},
      \clegend{Limpervioussurface}{Impervious surface}}
  \label{fig:Dx}
\end{figure}

Fig.\ref{fig:Dx} presents the visual comparison results of the proposed DepthSeg framework and several comparative methods on the LiuZhou dataset. The primary challenges in semantic segmentation for the LiuZhou dataset stem from issues such as densely packed buildings, varying road materials, and complex vegetation types, which often lead to land-cover misclassification. To comprehensively compare the strengths and weaknesses of the methods in different scenarios, urban areas and suburban regions are selected for detailed examination through magnified views.

In Fig.~\ref{fig:Dx}, the mapping results of the proposed DepthSeg in the test areas are closer to the ground truth compared to those of the baseline methods, demonstrating higher accuracy. Specifically, in the magnified results for densely populated urban areas, the baseline methods exhibit misclassification among roads, buildings, and impervious surfaces. By integrating depth information and leveraging the principle that different objects have distinct elevation characteristics, DepthSeg effectively distinguishes spectrally similar targets. In the suburban areas, the magnified results reveal issues in the baseline methods, such as road misclassification and blurred object boundaries. In contrast, DepthSeg achieves accurate land-cover classification and boundary extraction by combining 2D spectral-geometric features with 3D elevation-depth features, demonstrating clear advantages over the other methods. To provide a more objective evaluation of each method's performance on the LiuZhou dataset, a quantitative comparison is provided in Table~\ref{tab:Dl}.

\vspace{-0.4cm}
\begin{table}[H]
  \centering
  \caption{Accuracy assessment on the LiuZhou dataset.} 
    \begin{tabular}{ccccccc}
    \toprule
    \multirow{2}[4]{*}{Methods} & \multicolumn{6}{c}{Metrics(\%)} \\
\cmidrule{2-7}          & mPre  & mRecall & mF1   & mIoU  & OA    & Kappa \\
    \midrule
    Deeplabv3 & 57.49 & 54.64 & 54.96 & 42.97 & 79.95 & 72.33 \\
    ST-UNet & 69.97 & 63.89 & 63.52 & 50.57 & 81.36 & 74.55 \\
    CTCFNet & 70.64 & 72.25 & 71.12 & 58.19 & 86.37 & 81.31 \\
    DepthSeg-vits & 79.79 & \textcolor[rgb]{ 0,  0,  1}{80.78} & 80.17 & 68.99 & 90.21 & 86.54 \\
    DepthSeg-vitb & \textcolor[rgb]{ 0,  0,  1}{81.27} & 80.68 & \textcolor[rgb]{ 0,  0,  1}{80.85} & \textcolor[rgb]{ 0,  0,  1}{69.79} & \textcolor[rgb]{ 0,  0,  1}{90.67} & \textcolor[rgb]{ 0,  0,  1}{87.13} \\
    DepthSeg-vitl & \textcolor[rgb]{ 1,  0,  0}{81.78} & \textcolor[rgb]{ 1,  0,  0}{81.22} & \textcolor[rgb]{ 1,  0,  0}{81.42} & \textcolor[rgb]{ 1,  0,  0}{70.52} & \textcolor[rgb]{ 1,  0,  0}{91.05} & \textcolor[rgb]{ 1,  0,  0}{87.65} \\
    \bottomrule
    \end{tabular}%
  \label{tab:Dl}%
\end{table}%
\vspace{-0.3cm}

From Table~\ref{tab:Dl}, it can be seen that the proposed DepthSeg framework achieves high scores on the LiuZhou dataset. Among the different variants, DepthSeg-vitl records the highest scores across all the metrics in the LiuZhou dataset. Specifically, DepthSeg-vitl improves the Kappa and mIoU scores by 6.34\% and 12.33\%, compared to the best comparative method (CTCFNet). Moreover, all three proposed DepthSeg variants outperform the comparative methods across all the metrics. This can be attributed to the proposed DepthSeg framework's integration of depth information, which effectively mitigates the land cover misclassification caused by spectral confusion and shadow occlusion.

\section{Disscussion}

The accuracy improvements of DepthSeg primarily stem from the fine-tuning of the ViT encoder's lightweight adapter and the integration of the depth prompter. To assess the contribution of these two key components, an ablation study was conducted. The results of the ablation experiments are presented in Table~\ref{tab:Ablation}.

\vspace{-0.4cm}
\begin{table}[H]
  \tiny
  \centering
  \caption{Ablation study for the proposed modules in DepthSeg.} 
    \begin{tabular}{ccccccc}
    \toprule
    \multirow{2}[4]{*}{Encoder} & \multicolumn{2}{c}{Module} & \multicolumn{4}{c}{Metrics(\%)} \\
\cmidrule{2-7}          & \multicolumn{1}{p{4.19em}}{Adapter} & \multicolumn{1}{p{6.94em}}{Depth Prompter} & mF1   & mIoU  & OA    & Kappa \\
    \midrule
    \multicolumn{1}{c}{\multirow{4}[2]{*}{ViTs}} & ×     & ×     & 78.69 & 67.07 & 89.47 & 85.52 \\
          & √     & ×     & 78.86(\textcolor{blue}{+0.17}) & 67.34(\textcolor{blue}{+0.27}) & 89.65(\textcolor{blue}{+0.18}) & 85.77(\textcolor{blue}{+0.25}) \\
          & ×     & √     & 80.21(\textcolor{blue}{+1.52}) & 68.96(\textcolor{blue}{+1.89}) & 90.17(\textcolor{blue}{+0.70}) & 86.46(\textcolor{blue}{+0.94}) \\
          & √     & √     & 80.17(\textcolor{blue}{+1.48}) & 68.99(\textcolor{blue}{+1.92}) & 90.21(\textcolor{blue}{+0.74}) & 86.54(\textcolor{blue}{+1.02}) \\
    \midrule
    \multicolumn{1}{c}{\multirow{4}[2]{*}{ViTb}} & ×     & ×     & 79.66 & 68.23 & 90.03 & 86.26 \\
          & √     & ×     & 79.51(\textcolor{red}{-0.15}) & 68.13(\textcolor{red}{-0.1}) & 90.18(\textcolor{blue}{+0.15}) & 86.48(\textcolor{blue}{+0.22}) \\
          & ×     & √     & 80.49(\textcolor{blue}{+0.83}) & 69.31(\textcolor{blue}{+1.08}) & 90.4(\textcolor{blue}{+0.37}) & 86.79(\textcolor{blue}{+0.53}) \\
          & √     & √     & 80.85(\textcolor{blue}{+1.19}) & 69.79(\textcolor{blue}{+1.56}) & 90.67(\textcolor{blue}{+0.64}) & 87.13(\textcolor{blue}{+0.87}) \\
    \midrule
    \multicolumn{1}{c}{\multirow{4}[2]{*}{ViTl}} & ×     & ×     & 78.85 & 67.12 & 89.41 & 85.45 \\
          & √     & ×     & 79.07(\textcolor{blue}{+0.22}) & 67.51(\textcolor{blue}{+0.39}) & 89.67(\textcolor{blue}{+0.26}) & 85.81(\textcolor{blue}{+0.36}) \\
          & ×     & √     & 81.21(\textcolor{blue}{+2.36}) & 70.24(\textcolor{blue}{+3.12}) & 90.85(\textcolor{blue}{+1.44}) & 87.36(\textcolor{blue}{+1.91}) \\
          & √     & √     & 81.42(\textcolor{blue}{+2.57}) & 70.52(\textcolor{blue}{+3.40}) & 91.05(\textcolor{blue}{+1.64}) & 87.65(\textcolor{blue}{+2.20}) \\
    \bottomrule
    \end{tabular}%
  \label{tab:Ablation}%
\end{table}%
\vspace{-0.3cm}

Table~\ref{tab:Ablation} demonstrates that the combination of lightweight adapter and depth prompter enhances the performance of the DepthSeg framework across all three encoder types. In the DepthSeg framework using the ViT-l encoder, the lightweight adapter improves the baseline by 0.36\% in Kappa and 0.39\% in mIoU through fine-tuning the encoder's feature outputs. Meanwhile, the depth prompter achieves increases of 1.91\% in Kappa and 3.12\% in mIoU, due to the architectural enhancements. When combined, the lightweight adapter and depth prompter yield improvements of 2.20\% in Kappa and 3.40\% in mIoU, indicating that the primary benefit comes from embedding the depth information, which facilitates more accurate land-cover classification. In the ablation experiments with the ViT-B encoders, it can be observed that the lightweight adapter produces diminishing returns as the encoder parameter count increases. However, the combination of the lightweight adapter and depth prompter results in greater gains than the depth prompter alone. This suggests a risk of reduced accuracy when fine-tuning high-parameter encoders without using the depth prompter. Regardless of whether used independently or together, the depth prompter contributes positively to the accuracy of the semantic segmentation.

\section{Conclusion}
In this paper, a novel semantic segmentation framework incorporating coupled depth prompts (DepthSeg) has been proposed. This framework is designed to model depth and height information, addressing the land cover misclassification caused by spectral confusion and shadow occlusion. To effectively extract and represent features from 2D remote sensing images, a lightweight adapter based on a pre-trained ViT encoder is introduced, allowing for the cost-effective fine-tuning of a large-parameter model. A depth prompter is also proposed, explicitly modeling elevation and depth features. Furthermore, a semantic classifier is designed to combine depth prompts with high-dimensional object features, enabling accurate land cover mapping. Experiments conducted on the LiuZhou dataset validated the accuracy and robustness of DepthSeg in various complex scenarios. In the future, we will further explore the underlying principles of various remote sensing semantic segmentation tasks. Joint semantic segmentation using multimodal data (such as optical, synthetic aperture radar, and hyperspectral data) will be investigated to support additional practical applications and needs.

\small
\bibliographystyle{IEEEtranN}
\balance
\bibliography{DepthSeg}

\begin{thebibliography}{17}
\providecommand{\natexlab}[1]{#1}
\providecommand{\url}[1]{#1}
\csname url@samestyle\endcsname
\providecommand{\newblock}{\relax}
\providecommand{\bibinfo}[2]{#2}
\providecommand{\BIBentrySTDinterwordspacing}{\spaceskip=0pt\relax}
\providecommand{\BIBentryALTinterwordstretchfactor}{4}
\providecommand{\BIBentryALTinterwordspacing}{\spaceskip=\fontdimen2\font plus
\BIBentryALTinterwordstretchfactor\fontdimen3\font minus
  \fontdimen4\font\relax}
\providecommand{\BIBforeignlanguage}[2]{{%
\expandafter\ifx\csname l@#1\endcsname\relax
\typeout{** WARNING: IEEEtranN.bst: No hyphenation pattern has been}%
\typeout{** loaded for the language `#1'. Using the pattern for}%
\typeout{** the default language instead.}%
\else
\language=\csname l@#1\endcsname
\fi
#2}}
\providecommand{\BIBdecl}{\relax}
\BIBdecl

\bibitem[Long et~al.(2015)Long, Shelhamer, and Darrell]{long2015fully}
J.~Long, E.~Shelhamer, and T.~Darrell, ``Fully convolutional networks for
  semantic segmentation,'' in \emph{Proceedings of the IEEE conference on
  computer vision and pattern recognition}, 2015, pp. 3431--3440.

\bibitem[Ronneberger et~al.(2015)Ronneberger, Fischer, and
  Brox]{ronneberger2015u}
O.~Ronneberger, P.~Fischer, and T.~Brox, ``U-net: Convolutional networks for
  biomedical image segmentation,'' in \emph{Med. Image Comput. Comput. Assist.
  Interv.}\hskip 1em plus 0.5em minus 0.4em\relax Springer, 2015, pp. 234--241.

\bibitem[Badrinarayanan et~al.(2017)Badrinarayanan, Kendall, and
  Cipolla]{badrinarayanan2017segnet}
V.~Badrinarayanan, A.~Kendall, and R.~Cipolla, ``Segnet: A deep convolutional
  encoder-decoder architecture for image segmentation,'' \emph{IEEE
  Transactions on Pattern Analysis and Machine Intelligence}, vol.~39, no.~12,
  pp. 2481--2495, 2017.

\bibitem[Chen et~al.(2017{\natexlab{a}})Chen, Papandreou, Kokkinos, Murphy, and
  Yuille]{chen2017deeplab}
L.-C. Chen, G.~Papandreou, I.~Kokkinos, K.~Murphy, and A.~L. Yuille, ``Deeplab:
  Semantic image segmentation with deep convolutional nets, atrous convolution,
  and fully connected crfs,'' \emph{IEEE Transactions on Pattern Analysis and
  Machine Intelligence}, vol.~40, no.~4, pp. 834--848, 2017.

\bibitem[Chen et~al.(2018)Chen, Zhu, Papandreou, Schroff, and
  Adam]{chen2018encoder}
L.-C. Chen, Y.~Zhu, G.~Papandreou, F.~Schroff, and H.~Adam, ``Encoder-decoder
  with atrous separable convolution for semantic image segmentation,'' in
  \emph{Proceedings of the European conference on computer vision (ECCV)},
  2018, pp. 801--818.

\bibitem[Zhao et~al.(2017)Zhao, Shi, Qi, Wang, and Jia]{zhao2017pyramid}
H.~Zhao, J.~Shi, X.~Qi, X.~Wang, and J.~Jia, ``Pyramid scene parsing network,''
  in \emph{Proceedings of the IEEE conference on computer vision and pattern
  recognition}, 2017, pp. 2881--2890.

\bibitem[Xie et~al.(2021)Xie, Wang, Yu, Anandkumar, Alvarez, and
  Luo]{xie2021segformer}
E.~Xie, W.~Wang, Z.~Yu, A.~Anandkumar, J.~M. Alvarez, and P.~Luo, ``Segformer:
  Simple and efficient design for semantic segmentation with transformers,'' in
  \emph{Proc. Advances in Neural Inf. Process. Syst.}, vol.~34, 2021.

\bibitem[He et~al.(2022)He, Zhou, Zhao, Zhang, Yao, and Xue]{ST-UNet}
X.~He, Y.~Zhou, J.~Zhao, D.~Zhang, R.~Yao, and Y.~Xue, ``Swin transformer
  embedding unet for remote sensing image semantic segmentation,'' \emph{IEEE
  Transactions on Geoscience and Remote Sensing}, vol.~60, pp. 1--15, 2022.

\bibitem[Lu et~al.(2024)Lu, Zhang, Du, Xu, and Liu]{CTCFNet}
C.~Lu, X.~Zhang, K.~Du, H.~Xu, and G.~Liu, ``Ctcfnet: Cnn-transformer
  complementary and fusion network for high-resolution remote sensing image
  semantic segmentation,'' \emph{IEEE Transactions on Geoscience and Remote
  Sensing}, vol.~62, pp. 1--17, 2024.

\bibitem[Ni et~al.(2015)Ni, Sun, Ranson, Pang, Zhang, and Yao]{NI2015}
W.~Ni, G.~Sun, K.~J. Ranson, Y.~Pang, Z.~Zhang, and W.~Yao, ``Extraction of
  ground surface elevation from zy-3 winter stereo imagery over deciduous
  forested areas,'' \emph{Remote Sensing of Environment}, vol. 159, pp.
  194--202, 2015.

\bibitem[Yu et~al.(2023)Yu, Ni, Liu, Zhao, Zhang, and Sun]{YU2023}
T.~Yu, W.~Ni, J.~Liu, R.~Zhao, Z.~Zhang, and G.~Sun, ``Extraction of tree
  heights in mountainous natural forests from uav leaf-on stereoscopic imagery
  based on approximation of ground surfaces,'' \emph{Remote Sensing of
  Environment}, vol. 293, p. 113613, 2023.

\bibitem[Chen et~al.(2023)Chen, Huang, Liu, Wang, Liu, Zhang, Su, and
  Zhang]{CHEN2023113802}
P.~Chen, H.~Huang, J.~Liu, J.~Wang, C.~Liu, N.~Zhang, M.~Su, and D.~Zhang,
  ``Leveraging chinese gaofen-7 imagery for high-resolution building height
  estimation in multiple cities,'' \emph{Remote Sensing of Environment}, vol.
  298, p. 113802, 2023.

\bibitem[Oquab et~al.(2024)Oquab, Darcet, Moutakanni, Vo, Szafraniec, Khalidov,
  Fernandez, Haziza, Massa, El-Nouby, Assran, Ballas, Galuba, Howes, Huang, Li,
  Misra, Rabbat, Sharma, Synnaeve, Xu, Jegou, Mairal, Labatut, Joulin, and
  Bojanowski]{oquab2024dinov2learningrobustvisual}
M.~Oquab, T.~Darcet, T.~Moutakanni, H.~Vo, M.~Szafraniec, V.~Khalidov,
  P.~Fernandez, D.~Haziza, F.~Massa, A.~El-Nouby, M.~Assran, N.~Ballas,
  W.~Galuba, R.~Howes, P.-Y. Huang, S.-W. Li, I.~Misra, M.~Rabbat, V.~Sharma,
  G.~Synnaeve, H.~Xu, H.~Jegou, J.~Mairal, P.~Labatut, A.~Joulin, and
  P.~Bojanowski, ``Dinov2: Learning robust visual features without
  supervision,'' 2024.

\bibitem[Ranftl et~al.(2021)Ranftl, Bochkovskiy, and Koltun]{dpt}
R.~Ranftl, A.~Bochkovskiy, and V.~Koltun, ``Vision transformers for dense
  prediction,'' in \emph{2021 IEEE/CVF International Conference on Computer
  Vision (ICCV)}, 2021, pp. 12\,159--12\,168.

\bibitem[Wang et~al.(2004)Wang, Bovik, Sheikh, and Simoncelli]{SSIM}
Z.~Wang, A.~Bovik, H.~Sheikh, and E.~Simoncelli, ``Image quality assessment:
  from error visibility to structural similarity,'' \emph{IEEE Transactions on
  Image Processing}, vol.~13, no.~4, pp. 600--612, 2004.

\bibitem[Yang et~al.(2024)Yang, Kang, Huang, Xu, Feng, and Zhao]{depthanything}
L.~Yang, B.~Kang, Z.~Huang, X.~Xu, J.~Feng, and H.~Zhao, ``Depth anything:
  Unleashing the power of large-scale unlabeled data,'' in \emph{2024 IEEE/CVF
  Conference on Computer Vision and Pattern Recognition (CVPR)}, 2024, pp.
  10\,371--10\,381.

\bibitem[Chen et~al.(2017{\natexlab{b}})Chen, Papandreou, Schroff, and
  Adam]{chen2017rethinking}
L.-C. Chen, G.~Papandreou, F.~Schroff, and H.~Adam, ``Rethinking atrous
  convolution for semantic image segmentation,'' \emph{arXiv preprint
  arXiv:1706.05587}, 2017.

\end{thebibliography}
\end{CJK}
\end{document}